# Linear-Time Outlier Detection via Sensitivity

**Mario Lucic**
ETH Zurich
lucic@inf.ethz.ch

**Olivier Bachem**
ETH Zurich
olivier.bachem@inf.ethz.ch

**Andreas Krause**
ETH Zurich
krausea@ethz.ch

## Abstract

Outliers are ubiquitous in modern data sets. Distance-based techniques are a popular non-parametric approach to outlier detection as they require no prior assumptions on the data generating distribution and are simple to implement. Scaling these techniques to massive data sets without sacrificing accuracy is a challenging task. We propose a novel algorithm based on the intuition that outliers have a significant influence on the quality of divergence-based clustering solutions. We propose *sensitivity* – the worst-case impact of a data point on the clustering objective – as a measure of outlierness. We then prove that *influence* – a (non-trivial) upper-bound on the sensitivity can be computed by a simple linear time algorithm. To scale beyond a single machine, we propose a communication efficient distributed algorithm. In an extensive experimental evaluation, we demonstrate the effectiveness and establish the statistical significance of the proposed approach. In particular, it outperforms the most popular distance-based approaches while being several orders of magnitude *faster*.

## 1 Introduction

"An outlying observation, or outlier, is one that appears to deviate markedly from other members of the sample in which it occurs" [Grubbs, 1969]. Outliers are ubiquitous in modern data sets. Due to noise, uncertainty and adversarial behavior, such observations are inherent to many real-world problems such as fraud detection, activity monitoring, intrusion detection and many others. Discriminating outliers from normal (inlier) data has been extensively studied both in statistics and machine learning [Hodge and Austin, 2004].

The classic parametric approach is to assume the underlying distribution (e.g., a Gaussian mixture), estimate the parameters on the data set, and discriminate the inliers from the outliers by means of a statistical test. Estimating parameters of such models can be computationally intensive, it requires a set of "normal" data for training, and it fails if the assumptions are invalidated [Papadimitriou et al., 2003]. An alternative, non-parametric technique is to embed the instances into a metric space and use distances between the instances as an indication of outlierness. The following generally accepted definition is due to Knorr et al. (2000) and generalizes several statistical outlier detection tests: An object $x$ in a data set $\mathcal{X}$ is an $(\alpha, \delta)$-outlier if at least a fraction $\alpha$ of the objects in $\mathcal{X}$ are at a distance greater than $\delta$ from $x$. The definition was extended by Ramaswamy et al. (2000) who propose the following definition: Given two integers, $k$ and $m$, an object $x \in \mathcal{X}$ is deemed an outlier if less than $m$ objects have a larger distance to the $k^{th}$ nearest neighbor than $x$. Finally, Angiulli and Pizzuti (2002) suggested yet another extension whereby the objects are ranked by the sum of distances from the $k$ nearest objects. These definitions imply that one can compute the outlierness score based entirely on pairwise distances. The popularity of these techniques lies in the fact that they are easy to implement and require no prior assumptions on the data generating distribution. As a result, distance-based methods are ubiquitous in supervised as well as unsupervised outlier detection.

The aforementioned methods are inherently non-scalable as they require the computation of pairwise distances between all input points. A traditional approach to scaling-up distance-based techniques is *indexing* whereby one creates a data structure that provides approximate answers to nearest-neighbor type queries [Liu et al., 2006; Arya et al., 1998]. An alternative approach is to compute the distances only with respect to some subset of the data set. The issue with the former technique is that the effectiveness of the data structure drastically reduces with the increase in dimensionality. For the latter techniques it is unclear how much accuracy is sacrificed for performance and how to reduce the variance of the estimated scores.

**Our contributions.** We provide a linear-time and space technique for outlier detection. Based on the intuition that outliers severely influence the result of (distance-based) clustering, we show that *sensitivity* – the worst case impact of a point on *all* possible clusterings – is an effective measure for the outlierness of a point. We show how to approximate this measure and prove that in general a better approximation is not possible. Furthermore, we show that the proposed method is robust with respect to the choice of the distance function. We demonstrate the effectiveness of the proposed algorithm by comparing it to a variety of state-of-the-art distance-based techniques on real-world as well as synthetic data sets.

## 2 Background and Related Work

We start by describing the framework used to compare the most popular distance-based outlier detection schemes. Given a metric space $(\mathcal{M}, \mathrm{d})$ and a set $\mathcal{X} = \{x_i\}_{i=1}^n \subseteq \mathcal{M}$, let $q : \mathcal{M} \to [0, +\infty)$ be a function that assigns an *outlierness* score to each $x \in \mathcal{X}$. The scoring function $q$ in conjunction with a threshold $\delta \in [0, 1]$ implies a binary classifier

$$f_{q,\delta}(x) = \begin{cases} 1 & q(x) \geq \delta \\ 0 & \text{otherwise.} \end{cases}$$

We proceed by parameterizing the most popular distance-based outlier detection algorithms.

**Nearest neighbors (KNN).** The most popular distance-based technique was developed by Knorr *et al.* (2000), wherein an object $x \in \mathcal{X}$ is deemed an $(\alpha, \delta)$-outlier if at least a fraction $\alpha$ of all objects are more than $\delta$ away from $x$. Formally, if

$$|\{x' \in \mathcal{X} \mid \mathrm{d}(x, x') > \delta\}| \geq \alpha n. \quad (1)$$

Motivated by the problem of finding a good setting for the parameters $\alpha$ and $\delta$, Ramaswamy *et al.* (2000) developed an algorithm based on $k$ nearest neighbors (KNN): outlierness score is equal to the distance to the $k$-th nearest neighbor, i.e.

$$q(x) := \mathrm{d}^k(x, \mathcal{X}),$$

where $\mathrm{d}(x, S) = \min_{y \in S} d(x, y)$. Setting $\alpha = (n-k)/n$ the set of outliers defined by Equation 1 is the set of outliers

$$\{x \in \mathcal{X} \mid q(x) \geq \delta\}.$$

**Density-based local outliers (LOF).** Breunig *et al.* (2000) propose an algorithm which compares the local density around a point with that of its neighbors. Intuitively, points with scores significantly larger than one are more likely to be outliers. Formally, let $\mathrm{d}^k(x, \mathcal{X})$ be the distance of $x$ to its $k$-th nearest neighbor and let $N_k(x)$ be the set of $k$ nearest neighbors of $x$. Reachability-distance is defined as $\mathrm{d}_r(x, y) = \max\{\mathrm{d}^k(y, \mathcal{X}), \mathrm{d}(x, y)\}$ and the *local reachability density* as $\mathrm{LRD}(x) = |N_k(x)|/\sum_{y \in N_k(x)} \mathrm{d}_r(x, y)$. Finally, the outlierness score is computed as

$$q(x) := \frac{\sum_{y \in N_k(x)} \mathrm{LRD}(y)}{|N_k(x)| \mathrm{LRD}(x)}.$$

**Clustering.** A natural idea is to first partition the points into clusters and then use the distance to the closest cluster center as the outlierness measure. A popular choice is to run $k$-means and use the squared Euclidean distance to the closest center as the measure of outlierness [Hodge and Austin, 2004]. Formally, given the cluster centers $S$, one defines

$$q(x) := \mathrm{d}(x, S).$$

**Iterative sampling.** To scale-up the nearest neighbors approach, Wu and Jermaine (2006) compute distances from each $x \in \mathcal{X}$ to a random subsample $S_x \subseteq \mathcal{X}$ (different for each $x$) and define

$$q(x) := \mathrm{d}^k(x, S_x).$$

**One-time sampling.** Instead of sampling a random subsample for each object, Sugiyama and Borgwardt (2013) propose sampling a random subsample $S \subset \mathcal{X}$ and define

$$q(x) := \mathrm{d}^k(x, S).$$

**Other methods.** In this paper we focus on scalable unsupervised distance-based techniques with no prior assumptions on the data generating distribution. Hence, we do not consider *One-class SVMs* [Schölkopf *et al.*, 1999] nor parametric models in which an underlying probability density is assumed (such as fitting mixtures of Gaussians).

**Discussion.** The main issue with KNN and LOF is the computational complexity of $\mathcal{O}(n^2 d)$, where $d$ is the dimension of the underlying metric space. Some speedups can be obtained for low dimensional data sets by using KD-trees (or Balltrees) to approximate the nearest neighbor queries [Liu *et al.*, 2006]. However, this approach can perform as bad as exhaustive search for higher dimensional data. Both iterative sampling and one-time sampling reduce the computational complexity to $\mathcal{O}(nds)$, where $s$ is the subsample size. However, it is unclear how much accuracy is traded for scalability. The main concern with the subsampling approaches is the "naive" nature of the sample. While it is argued that in expectation both algorithms should approximate KNN, it is critical to control the variance of the estimate. In fact, given an imbalanced data set, a random sample will with high probability contain only the samples from the large clusters. In the experimental section we show that this is indeed a valid concern. The main drawback of the CLUSTERING approach is that it completely disregards the density of points in each cluster which makes it extremely sensitive to imbalanced data.

## 3 Sensitivity as a Measure of Outlierness

The intuition behind our approach is that outliers have a large impact on the quality of distance-based clustering solutions [Charikar *et al.*, 2001; Hautamäki *et al.*, 2005]. Figure 1 illustrates this with an example. We argue that *sensitivity* – the worst-case impact of a data point over *all $k$-clusterings* – is a natural measure of outlierness.

In a $k$-clustering problem (e.g. $k$-means) the goal is to minimize some additively decomposable distortion measure, i.e.

$$\mathrm{cost}(\mathcal{X}, Q) = \frac{1}{|\mathcal{X}|} \sum_{x \in \mathcal{X}} f_Q(x)$$

where $Q \in \mathcal{Q}$ is a possible clustering solution with $|Q| = k$ and $f_Q(x)$ is the cost contribution of a single point $x \in \mathcal{X}$. The sensitivity of a point is defined as the maximal ratio between its contribution and the average contribution, i.e.

$$\sigma(x) = \sup_{Q \subseteq \mathcal{Q}} \frac{f_Q(x)}{\frac{1}{|\mathcal{X}|} \sum_{x' \in \mathcal{X}} f_Q(x')}.$$

This concept was introduced by Langberg and Schulman (2010) in the context of integration by weighted sampling. One of the most prominent applications of sensitivity is constructing *coresets* — small weighted subsets of the data set that approximate the cost function uniformly over all clusterings $Q \in \mathcal{Q}$ [Feldman and Langberg, 2011]. Intuitively, one wants to sample the points that have a big impact on the objective function. In contrast, we propose *influence* – an upper bound on the sensitivity – as a direct measure of outlierness.

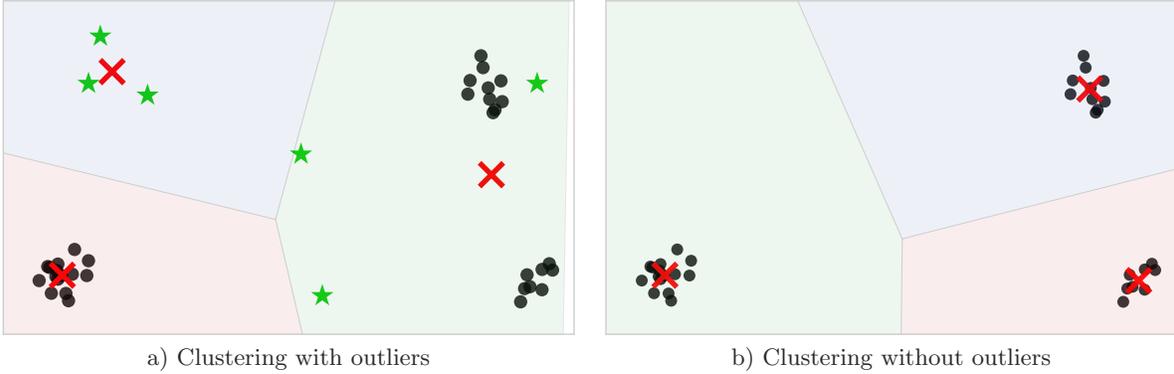

| a) Clustering with outliers | b) Clustering without outliers |

Figure 1: (a) Voronoi tessellation induced by a 3-means solution (red) on the data set (black) containing the outliers (green). (b) Optimal 3-means solution when outliers are removed. Even in this simple case, outliers have a significant impact.

Specifically, for $k$-means clustering, arguably the most popular distance-based $k$-clustering problem, we define

$$f_Q(x) = d(x, Q)^2 = \min_{q \in Q} \|x - q\|_2^2,$$

where $Q \in \mathbb{R}^{d \times k}$ and $x \in \mathcal{X} \subset \mathbb{R}^d$. The sensitivity of a point $x \in \mathcal{X}$ is thus

$$\sigma(x) = \sup_{Q \in \mathbb{R}^{d \times k}} \frac{d(x, Q)^2}{\frac{1}{|\mathcal{X}|} \sum_{x' \in \mathcal{X}} d(x', Q)^2}.$$

In general, $\sigma(x)$ cannot be efficiently computed as the supremum is with regards to all possible sets of $k$ points in $\mathbb{R}^d$.

**Influence.** We upper-bound the sensitivity $\sigma(x)$ by means of a surrogate function $s(x)$ uniformly over $x \in \mathcal{X}$ such that this bound is as tight as possible. To this end, let $s(x)$ be an upper-bound for $\sigma(x)$ and consider the difference between $s(x)$ and $\sigma(x)$. By definition,

$$\Delta = \frac{1}{|\mathcal{X}|} \sum_{x \in \mathcal{X}} |s(x) - \sigma(x)| = \frac{1}{|\mathcal{X}|} \sum_{x \in \mathcal{X}} s(x) - \frac{1}{|\mathcal{X}|} \sum_{x \in \mathcal{X}} \sigma(x).$$

Hence, obtaining a good approximation for $\sigma(x)$ is equivalent to minimizing

$$S = \frac{1}{|\mathcal{X}|} \sum_{x \in \mathcal{X}} s(x)$$

over all upper bounds $s(x)$. A naive approach is to use $s(x) = n$ which is valid by definition. However, such a trivial bound is useless for detecting outliers as it assigns the same outlierness score to all points. In what follows we illustrate how recent approaches to coreset construction can be used to efficiently obtain a much better, non-trivial bound.[1]

---
[1] Coresets are small summaries of the data set which guarantee that the solution on the summary is approximately the same as the solution on the original data set.

Lucic *et al.* (2016) propose a method that uses approximate clustering solutions to bound the sensitivity of each point. First, a rough approximation of the optimal solution is obtained by running the adaptive seeding step K-MEANS++ [Arthur and Vassilvitskii, 2007]. Such a solution $B = \{b_1, \ldots, b_k\}$ is already $\mathcal{O}(\log k)$ competitive with the optimal $k$-clustering solution.[2] The approximate clustering $B$ induces a Voronoi partitioning on the data set $\mathcal{X}$ whereby the points $x \in \mathcal{X}$ are assigned to their closest center in $B$. Based on this approximate solution, one can derive the following bound on the sensitivity:

$$s(x) = \underbrace{\frac{2\alpha\, d(x, b_x)^2}{\bar{c}_B}}_{(A)} + \underbrace{\frac{4\alpha \sum_{x' \in \mathcal{X}_x} d(x', b_x)^2}{|\mathcal{X}_x|\bar{c}_B}}_{(B)} + \underbrace{\frac{4n}{|\mathcal{X}_x|}}_{(C)} \quad (2)$$

where $\bar{c}_B = \frac{1}{n} \sum_{x' \in \mathcal{X}} d(x', B)^2$, $b_x$ is the closest center in $B$ to $x$ in terms of the squared Euclidean distance, and by $\mathcal{X}_x$ the set of all points $x' \in \mathcal{X}$ such that $b_x = b_{x'}$. Furthermore, if $\alpha = 16(\log_2 k + 2)$, $s(x)$ is a uniform upper bound for the influence (i.e. $\sigma(x) \leq s(x), \forall x \in \mathcal{X}$) and $S = \mathcal{O}(k)$.

We show that, for general data sets, this bound cannot be improved by more than a constant.

**Theorem 1.** *There exists a data set $\mathcal{X}$ such that*

$$S = \frac{1}{|\mathcal{X}|} \sum_{x \in \mathcal{X}} \sigma(x) = \Omega(k).$$

*Proof.* Consider a data set of size $n$ partitioned into $k$ sets of $n/k$ data points such that all points in a single set are located at the same location, but points in different sets are well separated. Now fix one of the sets and consider the clustering solution where one cluster center is situated $\sqrt{\epsilon} > 0$ from the location of this set, and $k - 1$ cluster centers are located at the points of the other sets. The average cost of this solution is $\epsilon/k$. Hence, for all points of this set the influence $\sigma(x)$ is lower bounded by $k$. Since the choice of the set was arbitrary and $\epsilon$ is a constant, it follows that $k$ is a uniform lower bound for the influence $\sigma(x)$. □

---
[2] Under natural assumptions on the data, a bicriteria approximation can even be computed in sublinear time [Bachem *et al.*, 2016].

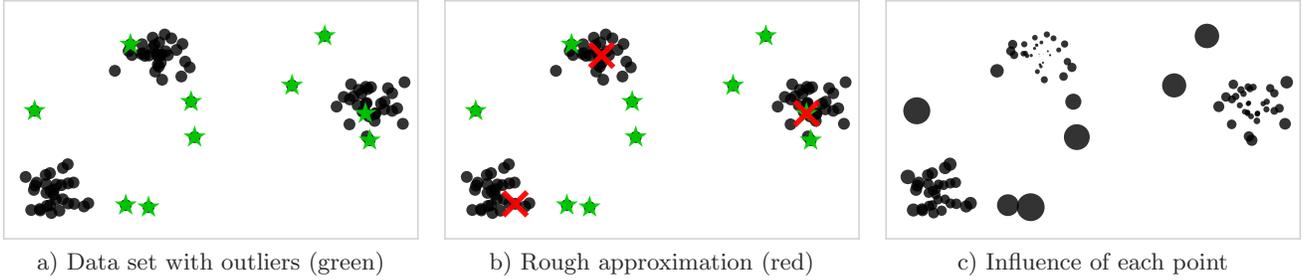

a) Data set with outliers (green)   b) Rough approximation (red)   c) Influence of each point

Figure 2: (a) Synthetic data set consisting of three clusters (black) and uniformly placed outliers (green). (b) Rough approximation (red). (c) The size of the points is proportional to the influence. Note that the influence increases when we move further away from the approximate solution or to regions with lower point density. One clearly recognizes the outlying points.

**Natural interpretation.** In the context of outlier detection, *influence* defined by Equation 2 has a rather natural interpretation. In fact, the terms (A)-(C) correspond to:

(A) **Local outlierness**: Outlierness increases with the distance from the cluster mean.
(B) **Cluster spread**: Outlierness of points belonging to clusters with larger variance increases.
(C) **Inverse cluster density**: Outlierness decreases if the cluster is more dense.

Hence, the notion of influence generalizes several outlier detection methods. By considering only the terms (A) and (C), we recognize a variant of the reference-based outlier detection method by Pei *et al.* (2006) where both density and distances are considered. Considering only the term (C) reduces to detecting low density clusters. Considering only the term (A) yields the clustering-based approach.

**Generality and robustness.** In practice, the most used divergence measure is the squared Euclidean distance. If the underlying distribution is a multivariate Gaussian, this choice is justified. Lucic *et al.* (2016) show that the bound defined in Equation 2 is a valid upper bound for a general class of Bregman divergences (such as squared Euclidean, Mahalanobis and KL-divergence). Due to the close relationship between Bregman divergences and the exponential family mixtures, INFLUENCE is robust with respect to the choice of the divergence measure and, by extension, to the underlying distribution. From this perspective, outlierness of a point corresponds to its impact on the likelihood of the data, under the given regular exponential family mixture.

The model selection problem – choosing the right value of $k$ – is unavoidable. To this end we propose a robust variance reduction technique: compute the influence $s(x)$ for different values of $k$ and then average the resulting functions to produce a final measure of outlierness. INFLUENCE can be efficiently computed for various values of $k$ as the approximate clustering solutions can be computed for all values of $k$ in a single pass.

## 4 Efficient Computation of Influence

Building on the approach of Lucic *et al.* (2016), we compute a bound on the sensitivity of each point. The resulting algorithm is detailed in Algorithm 1 and the computational complexity is characterized by the following proposition.

**Algorithm 1** INFLUENCE
**Require:** $\mathcal{X}, k$
  Uniformly sample $x \in \mathcal{X}$ and set $B = \{x\}$.
  **for** $i \leftarrow 2, 3, \ldots, k$ **do**
    Sample $x \in \mathcal{X}$ with probability $\frac{\mathrm{d}(x,B)^2}{\sum_{x' \in \mathcal{X}} \mathrm{d}(x',B)^2}$.
    $B \leftarrow B \cup \{x\}$.
  $\alpha \leftarrow 16(\log_2 k + 2)$
  **for each** $b_i$ in $B$ **do**
    $P_i \leftarrow$ Set of points from $\mathcal{X}$ closest to $b_i$.
  $c_\phi \leftarrow \frac{1}{|\mathcal{X}|} \sum_{x' \in \mathcal{X}} \mathrm{d}(x', B)^2$
  **for each** $b_i \in B$ and $x \in P_i$ **do**
    $s(x) \leftarrow \frac{\alpha \, \mathrm{d}(x,B)^2}{c_\phi} + \frac{2\alpha \sum_{x' \in P_i} \mathrm{d}(x',B)^2}{|P_i| c_\phi} + \frac{4|\mathcal{X}|}{|P_i|}$
  **return** $s$

**Proposition 1.** *Let $\mathcal{X} \subset \mathbb{R}^d$ and $k$ be a positive integer. Algorithm 1 calculates the influence $s(x)$ for $x \in \mathcal{X}$ as defined in Equation 2 in time $\mathcal{O}(nkd)$ and $\mathcal{O}(nd)$ space.*

**Distributed computation.** To scale beyond a single machine we provide a communication efficient algorithm that computes the influence based on a divide-and-conquer technique. We focus on the setting in which we have access to $m$ worker machines and one master machine that is used to synchronize the workers (equivalently, one might have access to a large machine with multiple cores). To obtain an approximate clustering, we use the `k-means∥` algorithm Bahmani *et al.* (2012) which distributes the computation of `k-means++`, while retaining the same theoretical guarantees. The remaining challenge is to compute the influence which depends not only on the local information (distance to $B$), but also on on the cluster densities and spread. These quantities can be computed efficiently with minimal overhead as detailed in Algorithm 2. The following proposition bounds the computational and communication complexity.

**Proposition 2.** *Let $\mathcal{X} \subset \mathbb{R}^d$, $m$ and $k$ be positive integers. Algorithm 2 computes the influence $s(x)$ for $x \in \mathcal{X}$ as defined in Equation 2 in time $\mathcal{O}(nkd(\log n)/m)$ with additional communication cost of $\mathcal{O}(n + k)$.*

|  | EXHAUSTIVE | | | | LINEAR | | | |
|---|---|---|---|---|---|---|---|---|
|  | KNN | KNN KD | LOF | LOF TREE | CLUSTER | ITERATIVE | ONETIME | INFLUENCE |
| IONOSPHERE | 0.927 | 0.927 | 0.842 | 0.842 | 0.877 ± 0.025 | 0.552 ± 0.033 | 0.846 ± 0.049 | **0.952 ± 0.008** |
| MFEAT | **0.679** | **0.679** | 0.498 | 0.498 | 0.503 ± 0.111 | 0.286 ± 0.053 | 0.573 ± 0.103 | 0.666 ± 0.038 |
| ARRHYTHMIA | 0.709 | 0.709 | 0.656 | 0.656 | 0.674 ± 0.012 | 0.601 ± 0.021 | 0.692 ± 0.011 | **0.722 ± 0.009** |
| WDBC | 0.612 | 0.612 | 0.425 | 0.425 | 0.554 ± 0.034 | 0.489 ± 0.022 | 0.597 ± 0.059 | **0.649 ± 0.011** |
| SEGMENTATION | 0.475 | 0.475 | 0.448 | 0.448 | 0.423 ± 0.194 | 0.138 ± 0.042 | 0.367 ± 0.099 | **0.514 ± 0.046** |
| PIMA | 0.514 | 0.514 | 0.403 | 0.403 | 0.464 ± 0.009 | 0.411 ± 0.014 | 0.485 ± 0.033 | **0.541 ± 0.006** |
| GAUSS-1K | 1.000 | 1.000 | 1.000 | 1.000 | 0.814 ± 0.054 | 0.018 ± 0.000 | 0.956 ± 0.131 | **1.000 ± 0.000** |
| OPTDIGITS | 0.189 | 0.189 | 0.158 | 0.158 | **0.192 ± 0.040** | 0.072 ± 0.012 | 0.149 ± 0.022 | 0.188 ± 0.013 |
| MNIST-2D | 0.127 | 0.127 | 0.060 | 0.060 | 0.117 ± 0.036 | 0.022 ± 0.020 | 0.087 ± 0.019 | **0.159 ± 0.015** |
| SPAMBASE | 0.418 | 0.418 | **0.428** | 0.428 | 0.404 ± 0.013 | 0.391 ± 0.005 | 0.411 ± 0.015 | 0.424 ± 0.004 |
| KDDCUP-100K | - | 0.426 | - | 0.579 | **0.732 ± 0.050** | 0.404 ± 0.002 | 0.695 ± 0.057 | 0.712 ± 0.007 |
| GAUSS-1M | - | 1.000 | - | 0.500 | 1.000 ± 0.000 | 0.000 ± 0.000 | 0.942 ± 0.231 | **1.000 ± 0.000** |
| KDDCUP-FULL | - | - | - | - | 0.355 ± 0.037 | 0.204 ± 0.021 | 0.291 ± 0.163 | **0.642 ± 0.016** |
| AVERAGE | 0.565 | 0.590 | 0.492 | 0.500 | 0.547 | 0.276 | 0.546 | **0.628** |
| AVG. RANK | 3.900 | 2.583 | 6.800 | 5.333 | 4.077 | 7.077 | 4.769 | **1.615** |
| RMSD | 0.023 | 0.091 | 0.116 | 0.184 | 0.118 | 0.465 | 0.120 | **0.007** |

Table 1: Performance measured by AUPRC on both real-world and synthetic data sets. The influence-based algorithm outperforms the competing linear-time as well as quadratic-time algorithms. We note that the average score and RMSD for the exhaustive methods presented in the table are optimistic as we ignore the instances for which the method failed to run.

---

**Algorithm 2** DISTRIBUTED INFLUENCE

**Require:** $\mathcal{X}, m, k, \alpha$
  Partition $\mathcal{X} \leftarrow \{\mathcal{X}_1, \ldots, \mathcal{X}_m\}$ to $m$ workers.
  Run $k$-means|| and communicate solution $B$ to all workers.
  **for all** $j \leftarrow 1, \ldots, m$ **in parallel do**
    **for each** $b_i$ in $B$ **do**
      $P_{ij} \leftarrow$ Points from $\mathcal{X}_j$ closest to $b_i$.
      $n_{ij} \leftarrow |P_{ij}|, e_{ij} \leftarrow \sum_{p \in P_{ij}} d(p, b_i)$
    Communicate $\{(n_{ij}, e_{ij})\}_{i=1}^k$ to master.
  **for all** $i \leftarrow 1, \ldots, k$ **do**
    $n_i^\star \leftarrow \sum_{j=1}^m n_{ij}, e_i^\star \leftarrow \sum_{j=1}^m e_{ij}$
  Communicate $\{(n_i^\star, e_i^\star)\}_{i=1}^k$ and $|\mathcal{X}|$ to all workers.
  **for all** $j \leftarrow 1, \ldots, m$ **in parallel do**
    **for all** $i \leftarrow 1, \ldots, k$ **and** $x \in P_{ij}$ **do**
      $s(x) \leftarrow \frac{\alpha |\mathcal{X}| d(x,B)^2}{\sum_{i=1}^k e_i^\star} + \frac{2\alpha |\mathcal{X}| e_i^\star}{n_i^\star \sum_{i=1}^k e_i^\star} + \frac{4|\mathcal{X}|}{n_i^\star}$
  **return** $s$

## 5 Experiments

In this section we demonstrate that influence is an effective measure of outlierness. We evaluate the proposed algorithm on both real-world and synthetic data sets.

**Evaluation measure.** Precision-recall curves illustrate the performance of a binary classifier as its discrimination threshold is varied. They are closely-related to receiver operating characteristic (ROC), but are more suited for imbalanced data sets [Goadrich *et al.*, 2006]. We use the area under the precision-recall curve (AUPRC) as a single number summary. Given an outlierness function $q$ and a threshold $\delta$ we construct the binary classifier defined in Section 2.

**Data sets.** The experimental evaluation is applied on a variety of real-world data sets available on UCI [Asuncion and Newman, 2007] as well as on synthetic data sets. As most of the data sets were intended as classification tasks we opted for the following setup commonly used in the literature:

1. All instances in the smallest class are outliers [Sugiyama and Borgwardt, 2013].
2. A subset of classes is deemed as inliers and three elements of the remaining classes are selected as outliers [Kriegel and Zimek, 2008; Hodge and Austin, 2004].

The relevant information is summarized in Table 2. The value "S" in the IN column implies the first strategy. Otherwise, it represents the inlier classes which were selected following Kriegel and Zimek (2008). For the synthetic data sets, we follow the approach from Sugiyama and Borgwardt (2013) and generate inliers from a Gaussian mixture model with 5 equally weighted components and 30 outliers from a uniform distribution in the range from the minimum to the maximum values of inliers. We standardize the features to zero mean and unit variance.

| DATA SET | N | D | IN |
|---|---|---|---|
| IONOSPHERE | 351 | 34 | S |
| MFEAT | 424 | 651 | 3, 9 |
| ARRHYTHMIA | 452 | 279 | S |
| WDBC | 569 | 30 | S |
| SEGMENTATION | 675 | 16 | 1, 2 |
| PIMA | 768 | 8 | S |
| GAUSS-1K | 1030 | 1000 | − |
| OPTDIGITS | 1155 | 64 | 3, 9 |
| MNIST-2D | 2852 | 2 | 3, 9 |
| SPAMBASE | 4601 | 57 | S |
| KDDCUP-100K | 100000 | 3 | S |
| GAUSS-1M | 1000030 | 20 | − |
| KDDCUP-FULL | 4898431 | 3 | S |

Table 2: Data set size, dimension, and inlier classes.

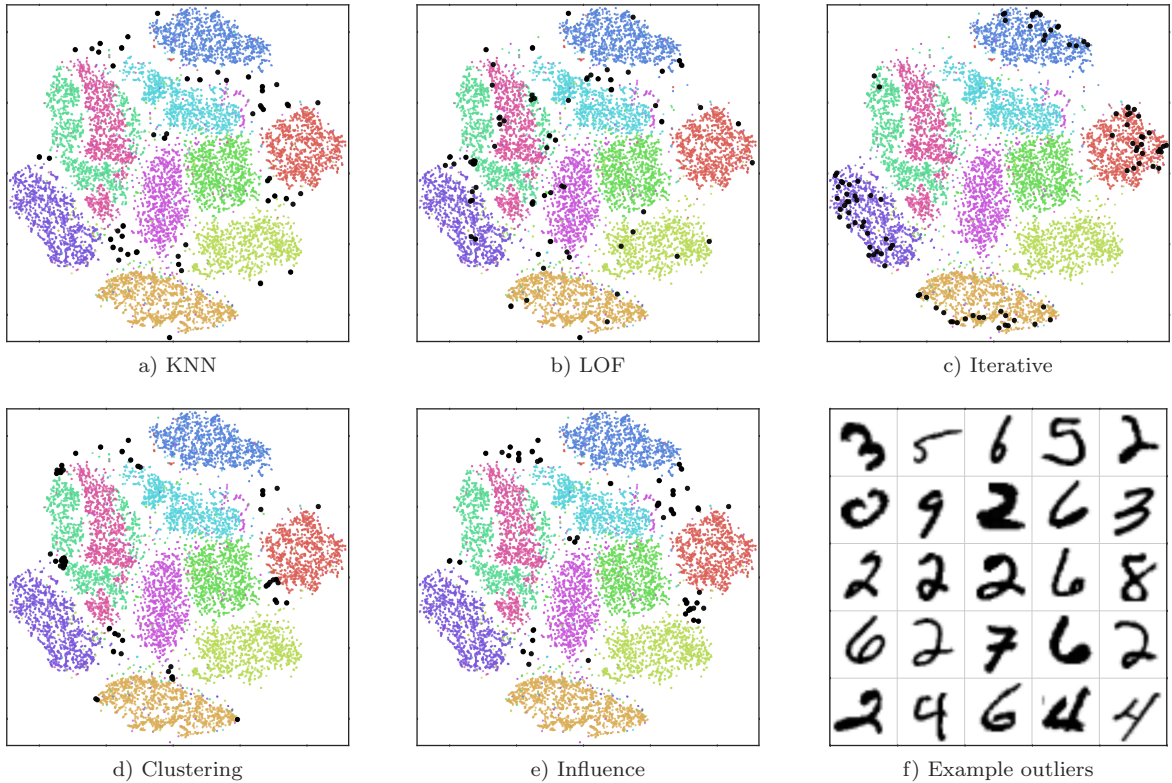

Figure 3: The MNIST data set embedded in 2D using t-SNE. The color clouds correspond to digits (e.g. red cloud corresponds to digit one) and the black points are the 64 points with the highest outlierness scores according to (a) KNN (b) LOF, (c) ITERATIVE SAMPLING, (d) CLUSTERING, (e) INFLUENCE. (f) Examples of outliers found by the proposed algorithm.

**Parameters.** We follow the parameter settings commonly used or suggested by the authors. For KNN and LOF we set $k = 10$ and $k = 5$, respectively [Bay and Schwabacher, 2003; Bhaduri *et al.*, 2011; Orair *et al.*, 2010]. For both ONE-TIME SAMPLING and ITERATIVE SAMPLING we set $s = 20$ and additionally $k = 5$ for ITERATIVE SAMPLING [Sugiyama and Borgwardt, 2013]. As our proposal, we apply Algorithm 1 with model averaging and $k \in \cup_{i=1}^{15}\{500/i\}$. For each algorithm with a random selection process we average 30 runs and we present the mean and variance of the AUPRC score.[3]

**Discussion.** Figure 3 provides some insight on outlier detection algorithms applied to the MNIST data set of handwritten digits. We embedded the data set in two dimensions using t-Distributed Stochastic Neighbor Embedding [Van Der Maaten, 2014] and assigned a color to each digit (e.g. the rightmost red region represents all instances of digit one). Due to the embedding, most outlying points lie between the point clusters. Clearly, KNN and INFLUENCE perform well and detect the between-cluster outliers. ITERATIVE SAMPLING tends to select points that are on the cluster boundary which implies that the chosen subsample is close to the mean of the data. The CLUSTERING approach is extremely sensitive to the initialization as well as the shape of the clusters. The LOF algorithm selects several outliers, but tends to overestimate the density which results in low performance on this data set.

The performance in terms of AUPRC is shown in Table 1. The proposed algorithm enjoys the lowest average root mean squared deviation (RMSD) from the best AUPRC among all methods. The AUPRC is statistically higher than that of any competing method (Mann-Whitney-Wilcoxon, $\alpha = 0.001$). Furthermore, it outperforms the baseline exhaustive methods (KNN and LOF) on 12 out of 13 data sets.

Running the exhaustive methods on larger data sets is not feasible due to the excessive computational and space requirements. To this end, we also compare against KNN with KD-TREES and LOF with BALL-TREES. On the GAUSS-1M data set we obtain a 27-fold speedup over the approximate KNN and 40-fold speedup over the approximate LOF. On KDDCUP-FULL the influence was computed in less than 10 minutes, while the computation time exceeds 24 hours for the exhaustive methods.

---

[3]The algorithms are implemented in Python 2.7 using NumPy and SciPy libraries and Cython for performance critical operations. The experiments were ran on Intel Xeon 3.3GHz machine with 36 cores and 1.5TB of RAM.

## Acknowledgments

This research was partially supported by ERC StG 307036 and the Zurich Information Security Center.

# References

[Angiulli and Pizzuti, 2002] Fabrizio Angiulli and Clara Pizzuti. Fast outlier detection in high dimensional spaces. In *PKDD*, volume 2, pages 15–26. Springer, 2002.

[Arthur and Vassilvitskii, 2007] David Arthur and Sergei Vassilvitskii. k-means++: The advantages of careful seeding. In *SODA*, pages 1027–1035. SIAM, 2007.

[Arya et al., 1998] Sunil Arya, David M Mount, Nathan S Netanyahu, Ruth Silverman, and Angela Y Wu. An optimal algorithm for approximate nearest neighbor searching fixed dimensions. *JACM*, 45(6):891–923, 1998.

[Asuncion and Newman, 2007] Arthur Asuncion and David Newman. UCI machine learning repository, 2007.

[Bachem et al., 2016] Olivier Bachem, Mario Lucic, S. Hamed Hassani, and Andreas Krause. Approximate k-means++ in sublinear time. In *AAAI*, February 2016.

[Bahmani et al., 2012] Bahman Bahmani, Benjamin Moseley, Andrea Vattani, Ravi Kumar, and Sergei Vassilvitskii. Scalable k-means++. *VLDB*, 5(7):622–633, 2012.

[Bay and Schwabacher, 2003] Stephen D Bay and Mark Schwabacher. Mining distance-based outliers in near linear time with randomization and a simple pruning rule. In *KDD*, pages 29–38. ACM, 2003.

[Bhaduri et al., 2011] Kanishka Bhaduri, Bryan L Matthews, and Chris R Giannella. Algorithms for speeding up distance-based outlier detection. In *KDD*, pages 859–867. ACM, 2011.

[Breunig et al., 2000] Markus M Breunig, Hans-Peter Kriegel, Raymond T Ng, and Jörg Sander. LOF: identifying density-based local outliers. In *SIGMOD Record*, volume 29, pages 93–104. ACM, 2000.

[Charikar et al., 2001] Moses Charikar, Samir Khuller, David M Mount, and Giri Narasimhan. Algorithms for facility location problems with outliers. In *SODA*, pages 642–651. SIAM, 2001.

[Feldman and Langberg, 2011] Dan Feldman and Michael Langberg. A unified framework for approximating and clustering data. In *STOC*, pages 569–578. ACM, 2011.

[Goadrich et al., 2006] Mark Goadrich, Louis Oliphant, and Jude Shavlik. Gleaner: Creating ensembles of first-order clauses to improve recall-precision curves. *Machine Learning*, 64(1-3):231–261, 2006.

[Grubbs, 1969] Frank E Grubbs. Procedures for detecting outlying observations in samples. *Technometrics*, 11(1):1–21, 1969.

[Hautamäki et al., 2005] Ville Hautamäki, Svetlana Cherednichenko, Ismo Kärkkäinen, Tomi Kinnunen, and Pasi Fränti. Improving k-means by outlier removal. In *Image Analysis*, pages 978–987. Springer, 2005.

[Hodge and Austin, 2004] Victoria J Hodge and Jim Austin. A survey of outlier detection methodologies. *Artificial Intelligence Review*, 22(2):85–126, 2004.

[Knorr et al., 2000] Edwin M Knorr, Raymond T Ng, and Vladimir Tucakov. Distance-based outliers: algorithms and applications. *VLDB*, 8(3-4):237–253, 2000.

[Kriegel and Zimek, 2008] Hans-Peter Kriegel and Arthur Zimek. Angle-based outlier detection in high-dimensional data. In *KDD*, pages 444–452. ACM, 2008.

[Langberg and Schulman, 2010] Michael Langberg and Leonard J Schulman. Universal $\varepsilon$-approximators for integrals. In *SODA*, pages 598–607. SIAM, 2010.

[Liu et al., 2006] Ting Liu, Andrew W Moore, and Alexander Gray. New algorithms for efficient high-dimensional nonparametric classification. *JMLR*, 7:1135–1158, 2006.

[Lucic et al., 2016] Mario Lucic, Olivier Bachem, and Andreas Krause. Strong coresets for hard and soft Bregman clustering with applications to exponential family mixtures. In *AISTATS*, 2016.

[Orair et al., 2010] Gustavo H Orair, Carlos HC Teixeira, Wagner Meira Jr, Ye Wang, and Srinivasan Parthasarathy. Distance-based outlier detection: consolidation and renewed bearing. *VLDB*, 3(1-2):1469–1480, 2010.

[Papadimitriou et al., 2003] Spiros Papadimitriou, Hiroyuki Kitagawa, Philip B Gibbons, and Christos Faloutsos. Loci: Fast outlier detection using the local correlation integral. In *ICDE*, pages 315–326. IEEE, 2003.

[Pei et al., 2006] Yaling Pei, Osmar R Zaiane, and Yong Gao. An efficient reference-based approach to outlier detection in large datasets. In *ICDM*, pages 478–487. IEEE, 2006.

[Ramaswamy et al., 2000] Sridhar Ramaswamy, Rajeev Rastogi, and Kyuseok Shim. Efficient algorithms for mining outliers from large data sets. In *SIGMOD Record*, volume 29, pages 427–438. ACM, 2000.

[Schölkopf et al., 1999] Bernhard Schölkopf, Robert C Williamson, Alex J Smola, John Shawe-Taylor, and John C Platt. Support vector method for novelty detection. In *NIPS*, volume 12, pages 582–588, 1999.

[Sugiyama and Borgwardt, 2013] Mahito Sugiyama and Karsten Borgwardt. Rapid distance-based outlier detection via sampling. In *NIPS*, pages 467–475, 2013.

[Van Der Maaten, 2014] Laurens Van Der Maaten. Accelerating t-SNE using tree-based algorithms. *JMLR*, 15(1):3221–3245, 2014.

[Wu and Jermaine, 2006] Mingxi Wu and Christopher Jermaine. Outlier detection by sampling with accuracy guarantees. In *KDD*, pages 767–772. ACM, 2006.